\documentclass{article}

\usepackage{arxiv}

\usepackage[utf8]{inputenc} 
\usepackage[T1]{fontenc}    
\usepackage{hyperref}
\hypersetup{colorlinks,allcolors=black}      
\usepackage{url}            
\usepackage{booktabs}       
\usepackage{amsfonts}       
\usepackage{nicefrac}       
\usepackage{microtype}      
\usepackage{lipsum}
\usepackage{times}
\usepackage{latexsym}
\usepackage{graphicx}
\usepackage{mathptmx}
\usepackage{amsmath}
\usepackage{caption}
\usepackage{multirow}
\usepackage{xspace}
\usepackage{amssymb}
\usepackage[ruled,vlined]{algorithm2e}

\usepackage{enumitem}

\title{Interpretable Multi-Head Self-Attention model for Sarcasm Detection in social media}

\author{
  Ramya Akula \\
  Department of Computer Science\\
  University of Central Florida, USA\\
  \texttt{ramya.akula@knights.ucf.edu} \\
   \And
  Ivan Garibay \\
  Department of Computer Science\\
  University of Central Florida, USA\\
  \texttt{igaribay@ucf.edu} \\
}

\begin{document}
\maketitle

\begin{abstract}
Sarcasm is a linguistic expression often used to communicate the opposite of what is said, usually something that is very unpleasant with an intention to insult or ridicule. Inherent ambiguity in sarcastic expressions, make sarcasm detection very difficult. In this work, we focus on detecting sarcasm in textual conversations from various social networking platforms and online media. To this end, we develop an interpretable deep learning model using multi-head self-attention and gated recurrent units. Multi-head self-attention module aids in identifying crucial sarcastic cue-words from the input, and the recurrent units learn long-range dependencies between these cue-words to better classify the input text. We show the effectiveness of our approach by achieving state-of-the-art results on multiple datasets from social networking platforms and online media. Models trained using our proposed approach are easily interpretable and enable identifying sarcastic cues in the input text which contribute to the final classification score. We visualize the learned attention weights on few sample input texts to showcase the effectiveness and interpretability of our model.
\end{abstract}


\section{Introduction}
Sarcasm is a rhetorical way of expressing dislike or negative emotions using exaggerated language constructs. It is an assortment of mockery and false politeness to intensify hostility without explicitly doing so. In face-to-face conversation, sarcasm can be identified effortlessly using facial expressions, gestures, and tone of the speaker. However, recognizing sarcasm in textual communication is not a trivial task as none of these cues are readily available. With the explosion of internet usage, sarcasm detection in online communications from social networking platforms \cite{ezaiza2016person, akula2019viztract}, discussion forums \cite{akula2019deepfork, akula2019forecasting}, and e-commerce websites has become crucial for opinion mining, sentiment analysis, and in identifying cyberbullies, online trolls. The topic of sarcasm received great interest from Neuropsychology \cite{shamay2005neuroanatomical} to Linguistics \cite{skalicky2018linguistic}, but developing computational models for automatic detection of sarcasm is still at its nascent phase. Earlier works on sarcasm detection on texts use lexical (content) and pragmatic (context) cues \cite{kreuz2007lexical} such as interjections, punctuation, and sentimental shifts, that are major indicators of sarcasm \cite{joshi2015harnessing}. In these works, the features are hand-crafted which cannot generalize in the presence of informal language and figurative slang widely used in online conversations.

With the advent of deep-learning, recent works \cite{ghosh2017magnets, ilic2018deep, ghosh2018sarcasm, xiong2019sarcasm, liu2019a2text}, leverage neural networks to learn both lexical and contextual features, eliminating the need for hand-crafted features. In these works, word embeddings are incorporated to train deep convolutional, recurrent, or attention-based neural networks to achieve state-of-the-art results on multiple large scale datasets. While deep learning-based approaches achieve impressive performance, they lack interpretability. In this work, we also focus on the interpretability of the model along with its high performance. The main contributions of our work are: 
\begin{itemize}
   \item Propose a novel, interpretable model for sarcasm detection using self-attention.
   \item Achieve state-of-the-art results on diverse datasets and exhibit the effectiveness of our model with extensive experimentation and ablation studies. 
   \item Exhibit the interpretability of our model by analyzing the learned attention maps. 
\end{itemize}

This paper is organized as follows: In Sections 2, and 3, we briefly mention the related works and describe our proposed multi-head self-attention architecture. Section 4 includes details on model implementation, experiments, datasets, and evaluation metrics. Performance and attention analysis of our model are described in Sections 5 and 6, followed by the conclusion of this work. 

\begin{figure*}
    \centering
    \includegraphics[scale=0.2] {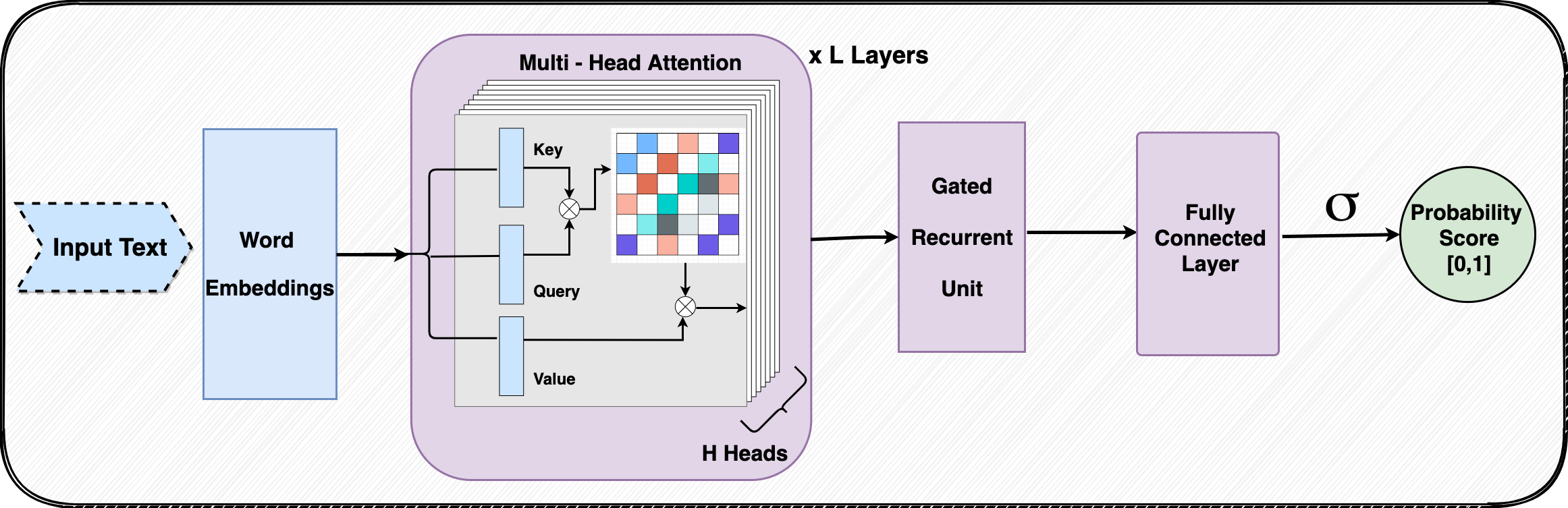}
    \caption{Multi head self-attention architecture for sarcasm detection. Pre-trained word embeddings are extracted for input text and are enhanced by an attention module with $L$ self-attention layers and $H$ heads per layer. Resultant features are passed through a Gated Recurrent Unit and a Feed-forward layer for classification.}
    \label{fig:Architecture} 
\end{figure*}

\section{Related Work}
Sarcasm is studied for many decades in social sciences, yet developing methods to automatically identify sarcasm in texts is a fairly new field of study. The state-of-the-art automated sarcasm detection models can be broadly segregated into content-based and context-based models. 

In content-based approaches, lexical and linguistic cues, syntactic patterns are used to train classifiers for sarcasm detection. \cite{carvalho2009clues, gonzalez2011identifying}, use linguistic features such as interjections, emoticons, and quotation marks. \cite{tsur2010icwsm, davidov2010semi} use syntactic patterns and lexical cues associated with sarcasm. The use of positive utterance in a negative context is used as a reliable feature to detect sarcasm by \cite{riloff2013sarcasm}. Linguistic features such as implicit and explicit context incongruity, are used by \cite{joshi2015harnessing}. In these works, only the input text is used to detect sarcasm without any context information. 

Context-based approaches increased in popularity in the recent past with the emergence of various online social networking platforms. As texts from these websites are prone to grammatical errors and extensive usage of slang, using context information helps in better identification of sarcasm. \cite{wallace2015sparse, poria2016deeper} detect sarcasm using sentiment and emotional information from the input text as contextual information. While, \cite{amir2016modelling, hazarika2018cascade} use personality features of the user as context, \cite{rajadesingan2015sarcasm,zhang2016tweet} use historical posts of the user to incorporate sarcastic tendencies. We show that, context information when available helps in improving the performance of the model but is not essential for sarcasm detection. 

Existing works by \cite{ptavcek2014sarcasm, wang2015twitter, wallace2015sparse, joshi2016harnessing}, use handcrafted features such as Bag of Words (BoW), Parts of Speech (POS), sentiment/emotions to train their classifiers. Other works by \cite{ghosh2016fracking, amir2016modelling, poria2016deeper, zhang2016tweet, vaswani2017attention, liu2019a2text} use deep-learning to learn meaningful features and classify them. The method which uses handcrafted features are easily interpretable but lack in performance. On the other hand, deep learning-based methods achieve high performance but lack interpretability. 

In our work, we propose a deep learning-based architecture for sarcasm detection which leverages self-attention to enable the interpretability of the model while achieving state-of-the-art performance on various datasets. 

\section{Proposed Approach}
Our proposed approach consists of five components: Data Pre-processing, Multi-Head Self-Attention, Gated Recurrent Units(GRU), Classification, and Model Interpretability. The architecture of our sarcasm detection model is shown in Figure \ref{fig:Architecture}. Data pre-processing involves converting input text to word embeddings, required for training a deep learning model. To this end, we first apply a standard tokenizer which does stop word removal, stemming, lemmatization, etc., to convert a sentence to a sequence of tokens. We employ pre-trained language models, to convert these tokens to word embeddings. These embeddings form the input to our multi-head self-attention module which identifies words in the input text that provide crucial cues for sarcasm. In the next step, the GRU layer aids in learning long-distance relationships among these highlighted words and output a single feature vector encoding the entire sequence. Finally, a fully-connected layer with sigmoid activation is used to get the final classification score.

\subsection{\textbf{Data Pre-processing}}
Word embeddings range from the clustering of words based on the local context to the embeddings based on a global context that considers the association between a word and every other word in a sentence. Most popular ones that rely on local context are Continuous Bag of Words (CBOW), Skip Grams \cite{mikolov2013efficient}, and Word2Vec \cite{mikolov2013distributed}. Other predictive models that capture global context are Global Vectors for word representation(GloVe) \cite{pennington2014glove}, FastText \cite{joulin2017bag}, Embeddings from Language Models (ELMO) \cite{peters2018deep} and Bidirectional Encoder Representations from Transformers (BERT) \cite{devlin2019bert}. In our work, we employ word embedding which captures global context as we believe it is essential for detecting sarcasm. We show the results of the proposed approach using multiple word embeddings, including, BERT, FastText, and GloVe.

\subsection{\textbf{Multi-Head Self-Attention}}
Given a sentence S, we apply standard tokenizer and use pre-trained models to obtain $D$ dimensional embeddings for individual words in the sentence. These embeddings $S$ = \{$e_1$, $e_2$, ..., $e_N$\}, $S \in \mathbb{R}^{N \times D}$ from the input to our model. To detect sarcasm in sentence S, it is crucial to identify specific words that provide essential cues such as sarcastic connotations and negative emotions. The importance of these cue-words is dependent on multiple factors based on different contexts. In our proposed model we leverage multi-head self-attention to identify these cue-words from the input text.

Attention is a mechanism to discover patterns in the input that are crucial for solving the given task. In deep learning, self-attention \cite{vaswani2017attention} is an attention mechanism for sequences, which helps in learning the task-specific relationship between different elements of a given sequence to produce a better sequence representation. In the self-attention module, three linear projections:
Key ($K$), Value ($V$), and Query ($Q$) of the given input sequence are generated, where $K, Q, V \in \mathbb{R}^{N \times D}$. Attention-map is computed based on the similarity between $K$, $Q$, and the output of this module $A \in \mathbb{R}^{N \times D}$ is the scaled dot-product between $V$ and the learned softmax attention ($QK^T$) as shown in Equation \ref{equ:self-attention}. 

\begin{equation}
    A = \text{softmax} \left( \frac{QK^T }{\sqrt{D}} \right)V
\label{equ:self-attention}
\end{equation}

In multi-head self-attention, multiple copies of the self-attention module are used in parallel. Each head captures different relationships between the words in the input text and identify those keywords that aid in classification. In our model, we use a series of multi-head self-attention layers (\#$L$) with multiple heads (\#$H$) in each layer. 

\subsection{\textbf{Gated Recurrent Units}}
Self-attention finds the words in the text which are important in detecting sarcasm. These words can be close to each other or farther apart in the input text. To learn long-distance relationships between these words, we use GRUs. These units are an improvement over standard recurrent neural networks and are designed to dynamically remember and forget the information flow using Reset ($r_t$) and Update ($z_t$) gates to solve the vanishing gradient problem.

In our model, we use a single layer of bi-directional GRU to process the sequence $A$, as these units makes use of the context information from both the directions. Given the input sequence $A \in \mathbb{R}^{N \times D}$, GRU computes hidden states $H$ = \{$h_1$, $h_2$, ..., $h_N$\},  $H \in \mathbb{R}^{N \times D}$ for every element in the sequence as follows:

\begin{equation}
\begin{split}
    r_t & = \sigma\left(W_rA_t + U_rh_{t-1} + b_r \right) \\
    z_t & = \sigma\left(W_zA_t + U_zh_{t-1} + b_z \right) \\
    \tilde{h}_t & = tanh\left(W_hA_t + U_h(r_{t} \odot h_{t-1})+ b_h \right) \\
    h_t  & = z_t \odot h_{t} + (1 - z_t) \odot \tilde{h}_{t-1} \\
\end{split}
\label{equ:gru}
\end{equation}
Here $\sigma$(.) is the element-wise sigmoid function and $W$, $U$, $b$ are the trainable weights and biases. $r_t$, $z_t$, $h_t$, $\tilde{h}_t$ $\in$ $\mathbb{R}^d$,
where $d$ is the size of the hidden dimension. We consider the final hidden state, $h_{N}$, which encodes all the information in the sequence, as an output from this module.

\subsection{\textbf{Classification}}
A single fully-connected feed-forward layer is used with sigmoid activation to compute the final output. Input to this layer is the feature vector $h_{N}$ from the GRU module and the output is a probability score $y \in [0, 1]$, computed as follows:

\begin{equation}
    y = \sigma\left(Wh_N + b\right),
\label{equ:fc1}
\end{equation}
where $W \in \mathbb{R}^{d \times 1}$ are the weights of this layer and $b$ is the bias term. Binary Cross Entropy (BCE) loss between the predicted output $y$ and the ground-truth label $\hat{y}$ is used to train the model.

\begin{equation}
    loss\left(y, \hat{y}\right) = \hat{y}log(y) + (1-\hat{y})log(1-y)
\label{equ:fc2}
\end{equation}
where $\hat{y} \in \{0, 1\}$ is the binary label i.e., 1:Sarcasm and 0:No-sarcasm.

\subsection{\textbf{Model Interpretability}} 
Developing models that can explain their predictions is crucial to building trust and faith in deep learning while enabling a wide range of applications with machine intelligence at its backbone. Existing deep learning network architectures such as convolutional and recurrent neural networks are not inherently interpretable and require additional visualization techniques \cite{zhou2016learning, selvaraju2017grad}. To avoid this, we in this work employ self-attention which is inherently interpretable and allows identifying elements in the input which are crucial for a given task.

\section{Experiments}
\subsection{\textbf{Datasets}}
\subsubsection{\textbf{Twitter \cite{riloff2013sarcasm}}}
In this dataset, the tweets that contain sarcasm are identified and labeled by the human annotators solely based on the contents of the tweets. These tweets do not depend on prior conversational context. Tweets with no sarcasm or that required prior conversational context are labeled as non-sarcastic.

\subsubsection{\textbf{Dialogues \cite{oraby2016creating}}}
This Sarcasm Corpus V2 Dialogues dataset is part of Internet Argument Corpus \cite{walker2012corpus} that includes annotated quote-response pairs for sarcasm detection. General sarcasm, hyperbole, rhetorical are the three categories in this dataset. In these quote-response pairs, a quote is a dialogic parent to the response. Therefore, a response post can be mapped to the same quote post or the post earlier in the thread. Here the quoted text is used as a context for sarcasm detection.

\subsubsection{\textbf{Twitter \cite{ghosh2017magnets}}}
In this dataset, tweets are collected using a Twitter bot named \textit{@onlinesarcasm}. This dataset not only contains tweets and replies to these tweets but also the mood of the user at the time of tweeting. The tweets/re-tweets of the users are the content and the replies to the tweets are the context.

\subsubsection{\textbf{Reddit \cite{khodak2018large}}} 
Self-annotated corpus for sarcasm, SARC 2.0 dataset contains comments from Reddit forums. Sarcastic comments by users are scrapped which are self-annotated by them using \textbackslash s token to indicate sarcastic intent. In our experiments, we use only the original comment without using any parent or child comments. "Main Balanced" and "Political" variants of the dataset are used in our experiments, the latter consists of comments only from the political subreddit.

\subsubsection{\textbf{Headlines \cite{misra2019sarcasm}}}
This news headlines dataset is collected from two news websites: Onion and Huffpost. The onion has sarcastic versions of current events whereas Huffpost has real news headlines. Headlines are used as content and the news article is used as context.

Details of these datasets, including the sample counts in train/test splits and the data source, are presented in Table \ref{tab:dataset-details}.

\begin{table}[h!]
\centering
\begin{tabular}{lccc}
    \hline 
    \hline
    \textbf{Source} & \textbf{Train} & \textbf{Test} & \textbf{Total} \\ 
    \hline
    \hline
    Twitter, 2013 & 1,368 & 588 & 1,956 \\
    Dialogues, 2016 & 3754 & 938 & 4,692 \\
    Twitter, 2017 & 51,189 & 3,742 & 54,931 \\
    Reddit, 2018 & 154,702 & 64,666 & 219,368 \\
    Headlines, 2019 & 22895 & 5724 & 28,619 \\
    \hline
    \hline
\end{tabular}
\caption{Statistics of datasets used in our experiments. Twitter, 2013 \cite{riloff2013sarcasm},
Dialogues, 2016 \cite{oraby2016creating},
Twitter, 2017 \cite{ghosh2017magnets},
Reddit, 2018 \cite{khodak2018large}, and
Headlines, 2019 \cite{misra2019sarcasm}. These are sourced from varied online platforms including social networks and discussion forums.}
\label{tab:dataset-details}
\end{table}

\subsection{\textbf{Implementation Details}}
We implement our model in PyTorch \cite{NEURIPS2019_9015}, a deep-learning framework in Python. To tokenize and extract word embeddings for the input text, we use publicly available resources \cite{Wolf2019HuggingFacesTS}. The embeddings for the words in the input text are passed through a series of multi-head self-attention layers $\#L$ , with multiple heads $\#H$ in each of the layers. The output from the self-attention layer is passed through a single bi-directional GRU layer with it's hidden dimension $d = 512$. The 512-dimensional output feature vector from the GRU layer is passed through the fully connected layer to get a 1-dimensional output. A sigmoid activation is applied to the final output and BCE loss is used to compute the loss between the ground truth and the predicted probability score. We use Adam optimizer to train our model with approximately 13 million parameters, using a learning rate of 1e-4, batch size of 64, and dropout set 0.2. We use one NVIDIA Pascal Titan-X with 16GB memory for all our experiments.  We set $\#H$ = 8 and $\#L$ = 3 in all our experiments for all the datasets.

\subsection{\textbf{Evaluation Metrics}}
We pose Sarcasm Detection as a classification problem, and use Precision, Recall, F1-Score, and Accuracy as evaluation metrics to test the performance of the trained models. \textit{Precision:} Ratio of the number of correctly predicted sarcastic sentences to the total number of predicted sarcastic sentences. \textit{Recall:} Ratio of correctly predicted sarcastic sentences to the actual number of sarcastic sentences in the ground-truth. \textit{F-score:} Harmonic mean of precision and recall. We use a threshold of 0.5 on the predictions from the model to compute these scores. Apart from these standard metrics we also compute Area Under the ROC Curve (AUC score) which is threshold independent.

\begin{table} [h!]
\centering
\begin{tabular}{lcccc}
\hline 
\hline
\textbf{Models} & \textbf{Precision} & \textbf{Recall} & \textbf{F1} & \textbf{AUC}\\ 
\hline
\hline
NBOW & 71.2 & 62.3 & 64.1 & - \\
Vanilla CNN & 71.0 & 67.1 & 68.5 & - \\
Vanilla LSTM & 67.3 & 67.2 & 67.2 & - \\
Attention LSTM & 68.7 & 68.6 & 68.7 & - \\
Bootstrapping \cite{riloff2013sarcasm} & 62.0 & 44.0 & 51.0 & - \\ EmotIDM \cite{farias2016irony} & - & - & 75.0 & - \\
Fracking Sarcasm \cite{ghosh2016fracking} & 88.3 & 87.9 & 88.1 & - \\
GRNN \cite{zhang2016tweet} & 66.3 & 64.7 & 65.4 & - \\
ELMo-BiLSTM \cite{ilic2018deep} & 75.9 & 75.0 & 75.9 & - \\
ELMo-BiLSTM FULL \cite{ilic2018deep} & 77.8 & 73.5 & 75.3 & -\\
ELMo-BiLSTM AUG \cite{ilic2018deep} & 68.4 & 70.8 & 69.4 & -\\
A2Text-Net \cite{liu2019a2text} & 91.7 & 91.0 & 90.0 & 97.0 \\
\hline
\hline
\multirow{2}{*}{\begin{tabular}[c]{@{}c@{}} \textbf{Our Model} \end{tabular}} & \textbf{97.9} & \textbf{99.6}  & \textbf{98.7} & \textbf{99.6}\\
& (+ 6.2 $\uparrow$) & (+ 8.6 $\uparrow$) & (+ 8.7 $\uparrow$) & (+ 2.6 $\uparrow$) \\
\hline
\hline
\end{tabular}
\caption{Results on Twitter dataset \cite{riloff2013sarcasm}.
} 
\label{tab:riloff}
\end{table}

\begin{table}[ht!]
\centering
\begin{tabular}{lcccc}
\hline 
\hline
\textbf{Models} & \textbf{Precision} & \textbf{Recall} & \textbf{F1} & \textbf{AUC}\\ 
\hline
\hline
Sarcasm Magnet 
\cite{ghosh2017magnets} & 73.3 & 71.7 & 72.5 & - \\ 
Sentence-level attention 
\cite{ghosh2018sarcasm} & 74.9 & 75.0 & 74.9 & - \\
Self Matching Networks 
\cite{xiong2019sarcasm} & 76.3 & 72.5 & 74.4 & - \\
A2Text-Net \cite{liu2019a2text} & 80.3 & 80.2 & 80.1 & 88.4 \\
\hline
\hline
\multirow{2}{*}{\begin{tabular}[c]{@{}c@{}} \textbf{Our Model} \end{tabular}} & \textbf{80.9} & \textbf{81.8} & \textbf{81.2} & \textbf{88.6} \\
 & (+ 0.6 $\uparrow$) & (+ 1.6 $\uparrow$) & (+ 1.1 $\uparrow$) & (+ 0.2 $\uparrow$) \\
\hline
\hline
\end{tabular}
\caption{Results on Twitter dataset \cite{ghosh2017magnets}.} 
\label{tab:magnets}
\end{table}

\begin{table}[h!]
\centering
\begin{tabular}{lcccc} 
\hline
\hline
\multirow{2}{*}{\textbf{Models}} & \multicolumn{2}{c}{\textbf{Main - Balanced}} & \multicolumn{2}{c}{\textbf{Political}} \\ \cline{2-5} & Accuracy & F1 & Accuracy & F1  \\ 
\hline
\hline
Bag-of-words & 63.0 & 64.0  & 59.0 & 60.0  \\ 
CNN & 65.0 & 66.0 & 62.0 & 63.0 \\ 
CNN-SVM \cite{poria2016deeper} & 68.0 & 68.0 & 70.65 & 67.0 \\ 
CUE-CNN \cite{amir2016modelling} & 70.0 & 69.0 & 69.0 & 70.0 \\ 
CASCADE \cite{hazarika2018cascade} & 77.0 & 77.0 & 74.0 & 75.0 \\ 
SARC 2.0 \cite{khodak2018large} & 75.0 & -  & 76.0 & - \\
ELMo-BiLSTM \cite{ilic2018deep} & 72.0 & -  & 78.0 & - \\
ELMo-BiLSTM FULL \cite{ilic2018deep} & 76.0 & 76.0  & 72.0 & 72.0 \\
\hline
\hline
\multirow{2}{*}{\begin{tabular}[c]{@{}c@{}} \textbf{Our Model} \end{tabular}} & \textbf{81.0} & \textbf{81.0} & \textbf{80.0} & \textbf{80.0} \\ 
& ( + 4.0 $\uparrow$) & ( + 4.0 $\uparrow$) &  ( + 2.0 $\uparrow$) & ( + 5.0 $\uparrow$) \\
\hline
\hline
\end{tabular}
\caption{Results on Reddit dataset SARC 2.0 and SARC 2.0 Political \cite{khodak2018large}.} 
\label{tab:sarc}
\end{table}

\begin{table}[ht]
\centering
\begin{tabular}{lcccc}
\hline
\hline
\textbf{Models} & \textbf{Precision} & \textbf{Recall} & \textbf{F1} & \textbf{AUC} \\
\hline
\hline
NBOW & 66.0 & 66.0 & 66.0 & - \\
Vanilla CNN & 68.4 & 68.1 & 68.2 & - \\
Vanilla LSTM & 68.3 & 63.9 & 60.7 & - \\
Attention LSTM & 70.0 & 69.6 & 69.6 & - \\
GRNN \cite{zhang2016tweet} & 62.2 & 61.8 & 61.2 & - \\
CNN-LSTM-DNN \cite{ghosh2016fracking} & 66.1 & 66.7 & 65.7 & - \\ 
SIARN \cite{tay2018reasoning} & 72.1 & 71.8 & 71.8 & - \\
MIARN \cite{tay2018reasoning} & 72.9 & 72.9 & 72.7 & - \\
ELMo-BiLSTM \cite{ilic2018deep} & 74.8 & 74.7 & 74.7 & - \\
ELMo-BiLSTM FULL \cite{ilic2018deep} & 76.0 & 76.0 & 76.0 & -\\
\hline
\hline
\multirow{2}{*}{\begin{tabular}[c]{@{}c@{}} \textbf{Our Model} \end{tabular}} & \textbf{77.4} & \textbf{77.2} & \textbf{77.2} & \textbf{0.834} \\
& ( + 1.2 $\uparrow$) & ( + 1.4 $\uparrow$) & ( + 1.2 $\uparrow$) & \\
\hline
\hline
\end{tabular}
\caption{Results on Sarcasm Corpus V2 Dialogues dataset \cite{oraby2016creating}
}
\label{tab:dialogues}
\end{table}

\begin{table}[h!]
\centering
\begin{tabular}{lccccc}
\hline 
\hline
\textbf{Models} & \textbf{Precision} & \textbf{Recall} & \textbf{F1} & \textbf{Accuracy} & \textbf{AUC} \\ 
\hline
\hline
Hybrid \cite{misra2019sarcasm} & - & - & - & 89.7 & - \\
A2Text-Net \cite{liu2019a2text} & 86.3 & 86.2 & 86.2 & - & 0.937 \\
\hline
\hline
\multirow{2}{*}{\begin{tabular}[c]{@{}c@{}} \textbf{Our Model} \end{tabular}} & \textbf{0.919} & \textbf{91.8} & \textbf{91.8} & \textbf{91.6} & \textbf{97.4} \\
& (+ 5.6 $\uparrow$) & ( + 5.6 $\uparrow$) & ( + 5.6 $\uparrow$) & ( + 1.9 $\uparrow$) & ( + 3.7 $\uparrow$) \\
\hline
\hline
\end{tabular}
\caption{Results on New Headlines dataset \cite{misra2019sarcasm}.} 
\label{tab:headlines}
\end{table}

\section{Results}
We present the results of our experiments on multiple publicly available datasets in this section. Results on Twitter datasets are presented in Table \ref{tab:riloff} and Table \ref{tab:magnets}. In the experiments with \cite{ghosh2017magnets} dataset, we do not use any additional information about the user or the context tweets. Hence, for a fair comparison, we present the results on this dataset under TTEA (Target Tweet Excluding Addressee) configuration. As evident from these tables, our multi-head self-attention model outperforms previous methods by a considerable margin. In Table \ref{tab:sarc}, we present the results on the Reddit SARC 2.0 dataset which is divided into two subsets: Main and Political. In both the datasets, our proposed approach outperforms previous methods.

Apart from Twitter and Reddit data we also experimented with data from other data sources such as Political Dialogues \cite{oraby2016creating} and News Headlines \cite{misra2019sarcasm}. In Table \ref{tab:dialogues}, we present results on Sarcasm Corpus V2 Dialogues dataset and in Table \ref{tab:headlines} we present result on News Headlines dataset. In both the datasets, we see considerable improvements. 

\subsection{\textbf{Ablation Study}}
Sarcasm Corpus V2 Dialogues dataset \cite{oraby2016creating} is used in the following experiments.

\subsubsection{\textbf{Ablation 1:}} 
We vary the number of self-attention layers and fix the number of heads per layer ($\#H$ = 8). From the results of this experiment presented in Table \ref{tab:ablationlayers}, we observe that as the number of self-attention layers increase ($\#L$ = 0, 1, 3, 5) the improvement in the performance of the model due to the additional layers saturate. Also, these results show that the proposed multi-head self-attention model achieves a 2\% improvement over the baseline model where only a single GRU layer is used without any self-attention layers. 

\begin{table}[h!]
\centering
\begin{tabular}{lccc}
\hline 
\hline
\textbf{\#$L$ - Layers} & \textbf{Precision} & \textbf{Recall} & \textbf{F1}\\
\hline
\hline
0 (GRU only) & 75.6 & 75.6 & 75.6 \\
1 Layer & 76.2 & 76.1 & 76.1 \\
3 Layers & 77.4 & 77.2 & 77.2 \\
5 Layers & 77.6 & 77.6 & 77.6 \\
\hline
\hline
\end{tabular}
\caption{Ablation study with varying number of attention layers \#$L$ and fixed Heads \textbf{\#$H$ = 8} on the Sarcasm Corpus V2 Dialogues dataset \cite{oraby2016creating}.}
\label{tab:ablationlayers}
\end{table}

\subsubsection{\textbf{Ablation 2:}} 
We vary the number of heads per layer with a fixed number of self-attention layers ($\#L$ = 3). Results of this experiments is presented Table \ref{tab:ablationheads}. We observe that the performance of the model also increases with the increase in the number of heads per self-attention layer.

\begin{table}[h!]
\centering
\begin{tabular}{cccc}
\hline 
\hline
\textbf{\#$H$ - Heads} & \textbf{Precision} & \textbf{Recall} & \textbf{F1}\\
\hline
\hline
1 Head & 74.9 & 74.5 & 74.4 \\
4 Heads & 76.9 & 76.8 & 76.8 \\
8 Heads & 77.4 & 77.2 & 77.2 \\
\hline
\hline
\end{tabular}
\caption{Ablation study with varying number of Heads \#$H$ and fixed Layers \textbf{\#$L$ = 3} on the Sarcasm Corpus V2 Dialogues dataset \cite{oraby2016creating}.}
\label{tab:ablationheads}
\end{table}

\subsubsection{\textbf{Ablation 3:}}
To further show the strength of our proposed network architecture, we perform one another ablation, in which we train our model with different word embedding such as Glove-6B, Glove-840B, ELMO, and FastText. Results are presented in Table \ref{tab:embedtypes}. These results show that the performance of our model is not due to the choice of word embeddings. With $\#H$ = 8 and $\#L$ = 3, the maximum possible batch size to train the model on 1 GPU with 16GB memory is 64. We set $\#H$ = 8 and $\#L$ = 3 in all our experiments for all the datasets. 

\begin{table}[h!]
\centering
\begin{tabular}{llcccc}
\hline 
\hline
\textbf{ Models} & \textbf{Embeddings} & \textbf{Precision} & \textbf{Recall} & \textbf{F1} & \textbf{AUC} \\ 
\hline
\hline
MIARN\cite{tay2018reasoning} & - & 72.9 & 72.9 & 72.7 & - \\
ELMo-BiLSTM FULL \cite{ilic2018deep} & ELMO & 76.0 & 76.0 & 76.0 & - \\
\hline
\hline
\multirow{5}{*}{\begin{tabular}[c]{@{}c@{}} Our Model \end{tabular}} 
& BERT & 77.4 & 77.2 & 77.2 & 83.4 \\
& ELMO & 76.7 & 76.7 & 76.7 & 80.8 \\
& FastText & 75.7 & 75.7 & 75.7 & 81.6 \\
& Glove 6B & 76.0 & 76.0 & 76.0 & 82.3 \\
& Glove 840B & 77.0 & 77.0 & 77.0 & 82.9 \\
\hline
\hline
\end{tabular}
\caption{Ablation study on various word embeddings on the Sarcasm Corpus V2 Dialogues dataset \cite{oraby2016creating}}
\label{tab:embedtypes}

\end{table}

\begin{figure}[h!]
\centering
    \includegraphics[scale=0.3] 
    {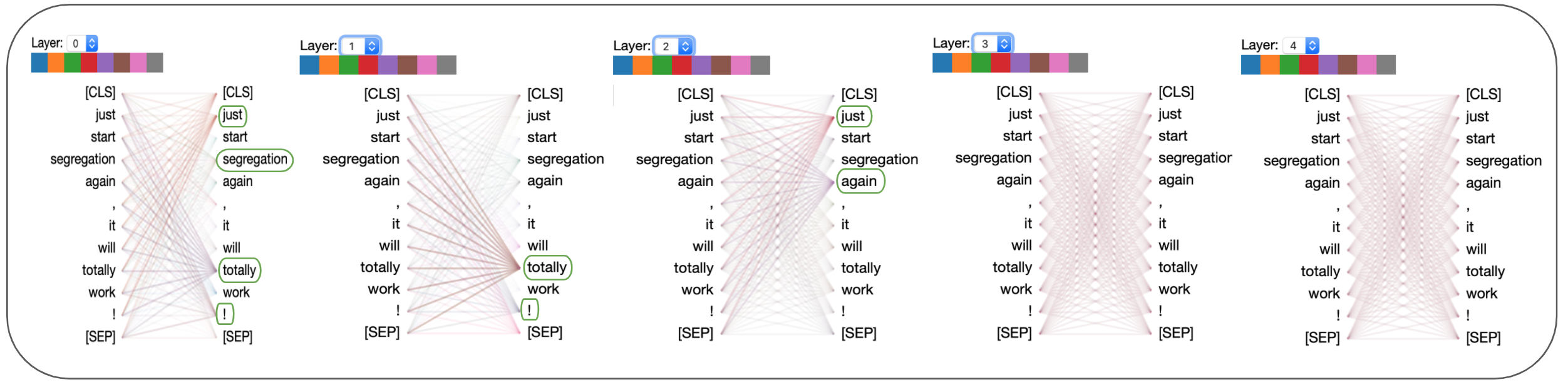}
    \caption{Attention analysis with sample sentence with sarcasm. Words providing cues for sarcasm, highlighted in green, are the words with higher attention weights.  The prediction score for this sentence by our model is 0.94.}
    \label{fig:HeadView_Sarc} 
\end{figure}

\begin{figure}[h!]
\centering
    \includegraphics[scale=0.3] {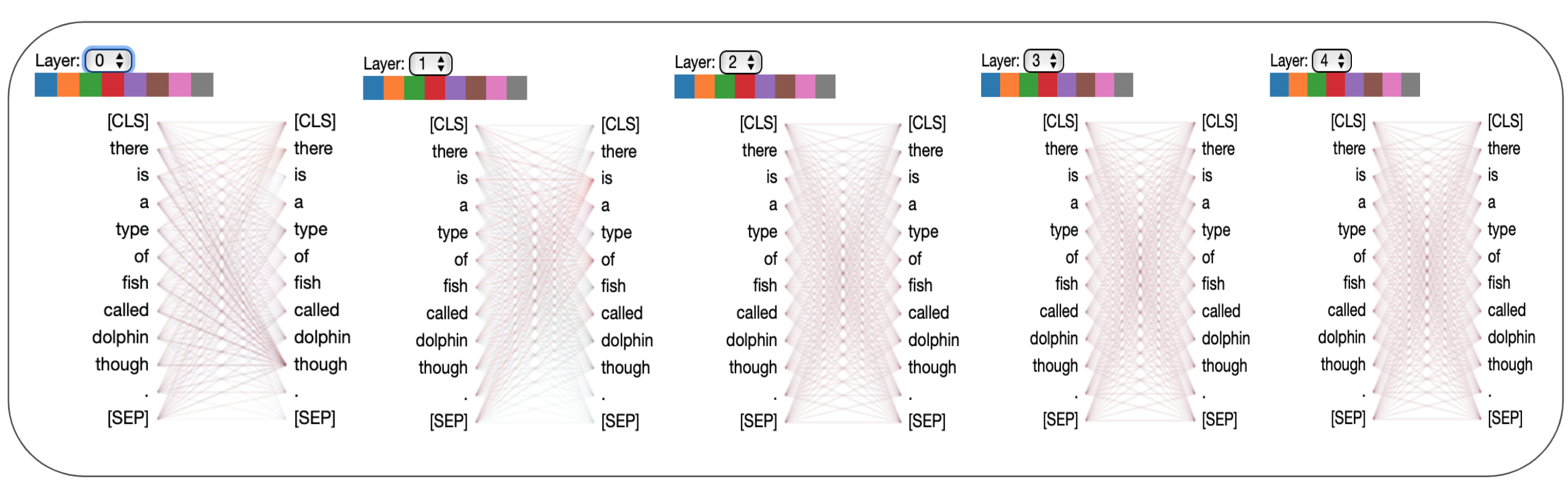}
    \caption{Attention analysis with sample sentence without sarcasm. Due to no presence of cues for sarcasm, every word in a sentence have similar attention weights. The prediction score for this sentence by our model is 0.15.}
    \label{fig:HeadView_NoSarc} 
\end{figure}

\begin{figure}[h!]
    \centering
    \includegraphics[scale=0.5] {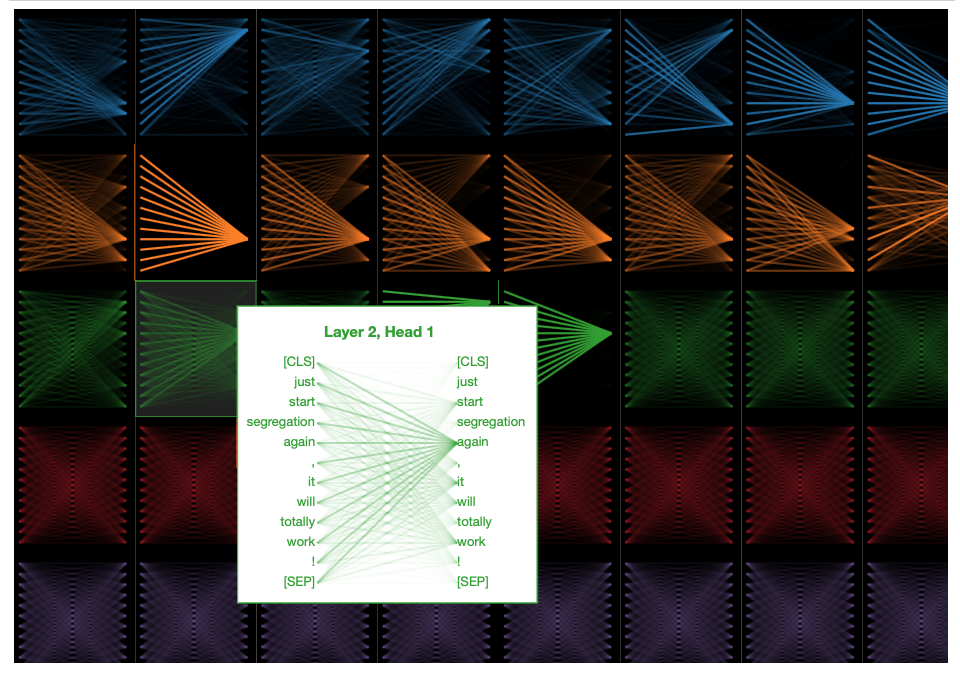}
    \caption{Attention analysis with sample sentence with sarcasm. Rows correspond to the different layers in the model and the columns correspond to the individual heads with a layer. When the input sentence contains sarcasm, we observe multiple heads, across layers attending to cue words in the input. }
    \label{fig:ModelView_Sarc} 
\end{figure}

\begin{figure}[h!]
    \centering
    \includegraphics[scale=0.5] {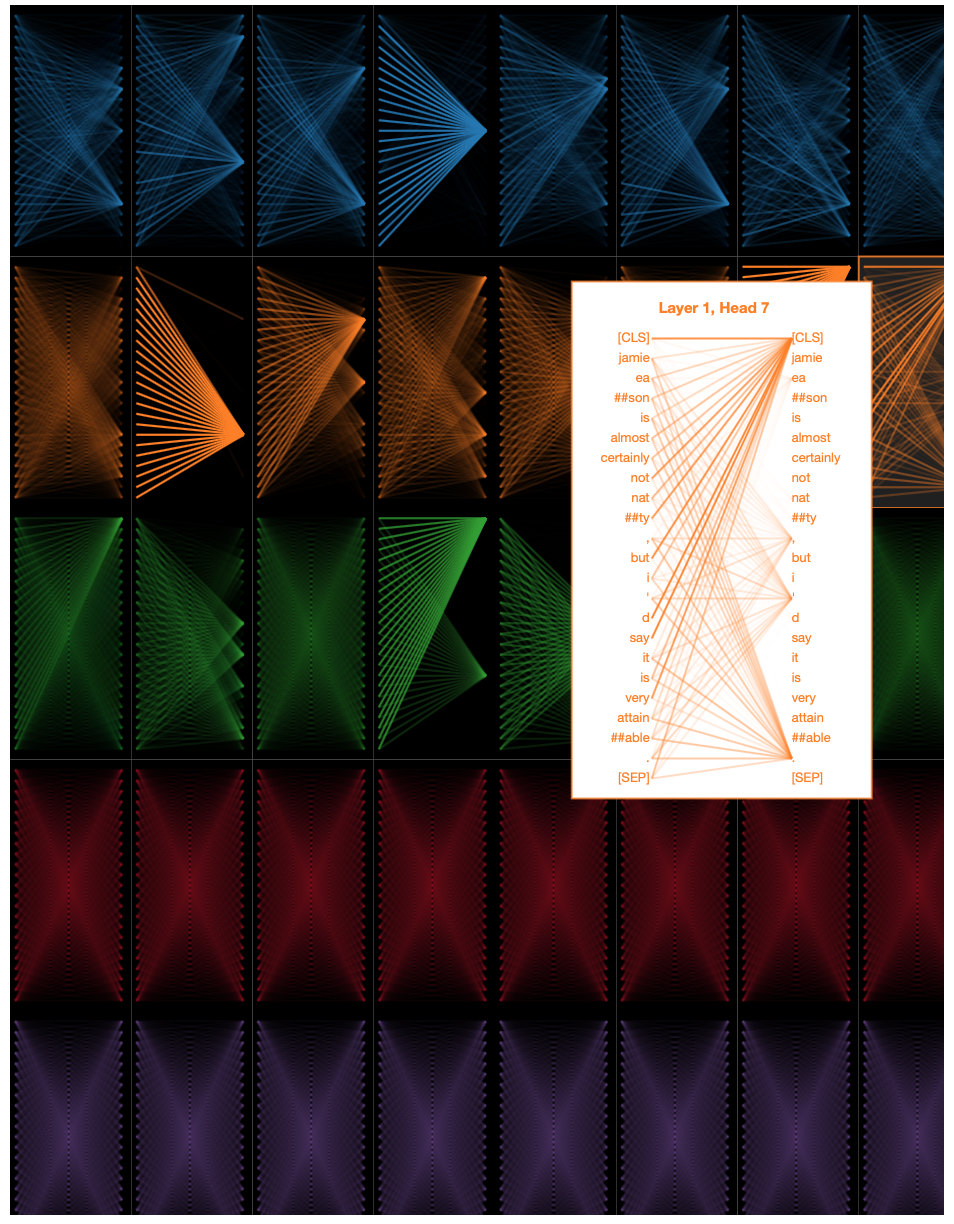}
    \caption{Attention analysis with sample sentence without sarcasm. Rows correspond to the different layers in the model and the columns correspond to the individual heads with a layer. When the input sentence contains no sarcasm, we observe that attention is distributed between multiple words in each head, across layers.}
    \label{fig:ModelView_NoSarc} 
\end{figure}

\begin{figure}[h!]
    \centering
    \includegraphics [scale=0.4] {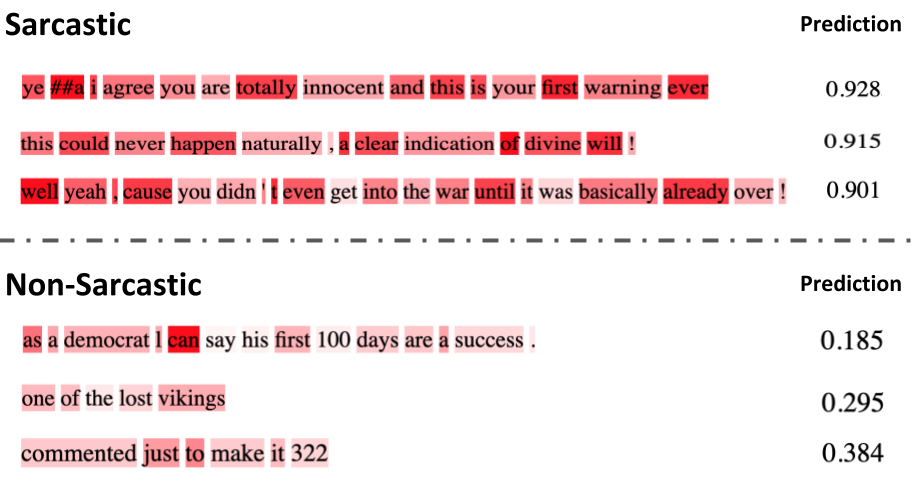}
    \caption{Visualization of the attention on individual words of sample sentences from both Sarcastic and Non-Sarcastic classes are shown in the column to the left. Probability scores predicted by our model are shown in the column to the right. High scores are predicted for sarcastic sentences and low scores for non-sarcastic sentences.}
    \label{fig:Analysis} 
\end{figure}

\section{Model Interpretability}
Attention maps from the individual heads of the self-attention layers provide the learned attention weights for each time-step in the input. In our case, each time-step is a word and we visualize the per-word attention weights for sample sentences with and without sarcasm from SARC 2.0 Main dataset. The model we used for this analysis has 5 attention layers with 8 heads per attention. Figures \ref{fig:HeadView_Sarc} and \ref{fig:HeadView_NoSarc} show attention analysis \cite{clark2019does} for sample sentences with and without sarcasm respectively. Each column in these figures corresponds to a single attention layer and attention weights between words in each head are represented using colored edges. The darkness of an edge indicates the strength of the attention weight. CLS and SEP are classifications and separator tokens from BERT.
Figures \ref{fig:ModelView_Sarc} and \ref{fig:ModelView_NoSarc}
are yet another visualization which provide a birds-eye view of attention across all the heads and layers in the model. Here rows correspond to 5 attention layers and the columns correspond to 8 heads in each layer. From both the visualizations, we observe that words receiving most attention vary between different heads in each layer and also across layers.

\subsection{\textbf{Attention Analysis}} 
For a sentence with sarcasm, Figure \ref{fig:HeadView_Sarc} shows that certain words receive more attention than others. For instance, words such as 'just', 'again', 'totally', '!', have darker edges connecting them with every other word in a sentence. These are the words in the sentence which hint at sarcasm and as expected these receive higher attention than others. Also, note that each cue word is attended by a different head in the first three layers of self-attention. In the final two layers, we observe that the attention is spread out to every word in the sentence indicating redundancy of these layers in the model. A sample sentence shown in Figure \ref{fig:HeadView_NoSarc} has no sarcasm, thus no word is highlighted by any head in any layer. In Figure \ref{fig:Analysis}, we visualize the distribution of attention over the words in a sentence for six sample sentences. Attention weight for a word is computed by first considering the maximum attention it receives across layers and then averaging the weights across multiple-heads in the layer. Finally, the weights for a word are averaged over all the words in the sentence. Stronger the highlight for a word, the higher is the attention weight placed on it by the model while classifying the sentence. Words from the sarcastic sentences with higher weights show that the model is able to detect sarcastic cues from the sentence. For example, the words "totally", "first", "ever" from the first sentence and "even", "until", "already" from the third sentence. These are the words that exhibit sarcasm in the sentences, which the model is able to successfully identify. In all the samples which are classified as non-sarcasm, the weights for the individual words are very low in comparison to cue-words from the sarcastic sentences. The probability of sarcasm predicted by our model for each of the sentences is shown on the right and their respective scores on the left column in Figure \ref{fig:Analysis}. Our model is able to predict a high score for sarcastic sentences and low scores for non-sarcastic sentences. 

\section{Conclusion}
In this work, we propose a novel multi-head self-attention based neural network architecture to detect sarcasm in a given sentence. Our proposed approach has 5 components: data pre-processing, multi-head self-attention module, gated recurrent unit module, classification, and model interpretability. Multi-head self-attention is used to highlight the parts of the sentence which provide crucial cues for sarcasm detection. GRUs aid in learning long-distance relationships among these highlighted words in the sentence. The output from this layer is passed through a fully-connected classification layer to get the final classification score. Experiments are conducted on multiple datasets from varied data sources and show significant improvement over the state-of-the-art models by all evaluation metrics. Results from ablation studies and analysis of the trained model are presented to show the importance of different components of our model. We analyze the learned attention weights to interpret our trained model and show that it can indeed identify words in the input text which provide cues for sarcasm.

\bibliographystyle{IEEEtran}
\bibliography{main}

\begin{thebibliography}{10}
\providecommand{\url}[1]{#1}
\csname url@samestyle\endcsname
\providecommand{\newblock}{\relax}
\providecommand{\bibinfo}[2]{#2}
\providecommand{\BIBentrySTDinterwordspacing}{\spaceskip=0pt\relax}
\providecommand{\BIBentryALTinterwordstretchfactor}{4}
\providecommand{\BIBentryALTinterwordspacing}{\spaceskip=\fontdimen2\font plus
\BIBentryALTinterwordstretchfactor\fontdimen3\font minus
  \fontdimen4\font\relax}
\providecommand{\BIBforeignlanguage}[2]{{%
\expandafter\ifx\csname l@#1\endcsname\relax
\typeout{** WARNING: IEEEtran.bst: No hyphenation pattern has been}%
\typeout{** loaded for the language `#1'. Using the pattern for}%
\typeout{** the default language instead.}%
\else
\language=\csname l@#1\endcsname
\fi
#2}}
\providecommand{\BIBdecl}{\relax}
\BIBdecl

\bibitem{ezaiza2016person}
H.~Ezaiza, S.~R. Humayoun, R.~AlTarawneh, and A.~Ebert, ``Person-vis:
  Visualizing personal social networks (ego networks),'' in \emph{Proceedings
  of the 2016 CHI Conference Extended Abstracts on Human Factors in Computing
  Systems}, 2016, pp. 1222--1228.

\bibitem{akula2019viztract}
R.~Akula and I.~Garibay, ``Viztract: Visualization of complex social networks
  for easy user perception,'' \emph{Big Data and Cognitive Computing}, vol.~3,
  no.~1, p.~17, 2019.

\bibitem{akula2019deepfork}
R.~Akula, N.~Yousefi, and I.~Garibay, ``Deepfork: Supervised prediction of
  information diffusion in github,'' \emph{Proceedings of the International
  Conference on Industrial Engineering and Operations Management}, 2019.

\bibitem{akula2019forecasting}
R.~Akula, Z.~Wieselthier, L.~Martin, and I.~Garibay, ``Forecasting the success
  of television series using machine learning,'' in \emph{2019
  SoutheastCon}.\hskip 1em plus 0.5em minus 0.4em\relax IEEE, 2019, pp. 1--8.

\bibitem{shamay2005neuroanatomical}
S.~G. Shamay-Tsoory, R.~Tomer, and J.~Aharon-Peretz, ``The neuroanatomical
  basis of understanding sarcasm and its relationship to social cognition.''
  \emph{Neuropsychology}, no.~3, p. 288, 2005.

\bibitem{skalicky2018linguistic}
S.~Skalicky and S.~Crossley, ``Linguistic features of sarcasm and metaphor
  production quality,'' in \emph{Proceedings of the Workshop on Figurative
  Language Processing}, 2018, pp. 7--16.

\bibitem{kreuz2007lexical}
R.~J. Kreuz and G.~M. Caucci, ``Lexical influences on the perception of
  sarcasm,'' in \emph{Proceedings of the Workshop on computational approaches
  to Figurative Language}.\hskip 1em plus 0.5em minus 0.4em\relax Association
  for Computational Linguistics, 2007, pp. 1--4.

\bibitem{joshi2015harnessing}
A.~Joshi, V.~Sharma, and P.~Bhattacharyya, ``Harnessing context incongruity for
  sarcasm detection,'' in \emph{Proceedings of the 53rd Annual Meeting of the
  ACL and the 7th IJCNLP}, 2015, pp. 757--762.

\bibitem{ghosh2017magnets}
A.~Ghosh and T.~Veale, ``Magnets for sarcasm: making sarcasm detection timely,
  contextual and very personal,'' in \emph{Proceedings of the 2017 Conference
  on EMNLP}, 2017, pp. 482--491.

\bibitem{ilic2018deep}
S.~Ilic, E.~Marrese-Taylor, J.~Balazs, and Y.~Matsuo, ``Deep contextualized
  word representations for detecting sarcasm and irony,'' in \emph{Proceedings
  of the 9th Workshop on Computational Approaches to Subjectivity, Sentiment
  and Social Media Analysis}, 2018, pp. 2--7.

\bibitem{ghosh2018sarcasm}
D.~Ghosh, A.~R. Fabbri, and S.~Muresan, ``Sarcasm analysis using conversation
  context,'' \emph{Computational Linguistics}, pp. 755--792, 2018.

\bibitem{xiong2019sarcasm}
T.~Xiong, P.~Zhang, H.~Zhu, and Y.~Yang, ``Sarcasm detection with self-matching
  networks and low-rank bilinear pooling,'' in \emph{The World Wide Web
  Conference}, 2019, pp. 2115--2124.

\bibitem{liu2019a2text}
L.~Liu, J.~L. Priestley, Y.~Zhou, H.~E. Ray, and M.~Han, ``A2text-net: A novel
  deep neural network for sarcasm detection,'' in \emph{2019 IEEE First
  International Conference on Cognitive Machine Intelligence (CogMI)}.\hskip
  1em plus 0.5em minus 0.4em\relax IEEE, 2019, pp. 118--126.

\bibitem{carvalho2009clues}
P.~Carvalho, L.~Sarmento, M.~J. Silva, and E.~De~Oliveira, ``Clues for
  detecting irony in user-generated contents: oh...!! it's so easy;-,'' in
  \emph{Proceedings of the 1st international CIKM workshop on Topic-sentiment
  analysis for mass opinion}, 2009, pp. 53--56.

\bibitem{gonzalez2011identifying}
R.~Gonz{\'a}lez-Ib{\'a}nez, S.~Muresan, and N.~Wacholder, ``Identifying sarcasm
  in twitter: a closer look,'' in \emph{Proceedings of the 49th Annual Meeting
  of the ACL: Human Language Technologies: Short Papers-Volume 2}, 2011, pp.
  581--586.

\bibitem{tsur2010icwsm}
O.~Tsur, D.~Davidov, and A.~Rappoport, ``Icwsm—a great catchy name:
  Semi-supervised recognition of sarcastic sentences in online product
  reviews,'' in \emph{Fourth International AAAI Conference on Weblogs and
  Social Media}, 2010.

\bibitem{davidov2010semi}
D.~Davidov, O.~Tsur, and A.~Rappoport, ``Semi-supervised recognition of
  sarcastic sentences in twitter and amazon,'' in \emph{Proceedings of the
  fourteenth conference on computational natural language learning}.\hskip 1em
  plus 0.5em minus 0.4em\relax Association for Computational Linguistics, 2010,
  pp. 107--116.

\bibitem{riloff2013sarcasm}
E.~Riloff, A.~Qadir, P.~Surve, L.~De~Silva, N.~Gilbert, and R.~Huang, ``Sarcasm
  as contrast between a positive sentiment and negative situation,'' in
  \emph{Proceedings of the 2013 Conference on EMNLP}, 2013, pp. 704--714.

\bibitem{wallace2015sparse}
B.~C. Wallace, E.~Charniak \emph{et~al.}, ``Sparse, contextually informed
  models for irony detection: Exploiting user communities, entities and
  sentiment,'' in \emph{Proceedings of the 53rd Annual Meeting of the ACL and
  the 7th IJCNLP}, 2015, pp. 1035--1044.

\bibitem{poria2016deeper}
S.~Poria, E.~Cambria, D.~Hazarika, and P.~Vij, ``A deeper look into sarcastic
  tweets using deep convolutional neural networks,'' in \emph{Proceedings of
  COLING 2016, the 26th International Conference on Computational Linguistics:
  Technical Papers}, 2016, pp. 1601--1612.

\bibitem{amir2016modelling}
S.~Amir, B.~C. Wallace, H.~Lyu, P.~Carvalho, and M.~J. Silva, ``Modelling
  context with user embeddings for sarcasm detection in social media,'' in
  \emph{Proceedings of The 20th SIGNLL Conference on Computational Natural
  Language Learning}, 2016, pp. 167--177.

\bibitem{hazarika2018cascade}
D.~Hazarika, S.~Poria, S.~Gorantla, E.~Cambria, R.~Zimmermann, and R.~Mihalcea,
  ``Cascade: Contextual sarcasm detection in online discussion forums,'' in
  \emph{Proceedings of the 27th International Conference on Computational
  Linguistics}, 2018, pp. 1837--1848.

\bibitem{rajadesingan2015sarcasm}
A.~Rajadesingan, R.~Zafarani, and H.~Liu, ``Sarcasm detection on twitter: A
  behavioral modeling approach,'' in \emph{Proceedings of the Eighth ACM
  International Conference on Web Search and Data Mining}, 2015, pp. 97--106.

\bibitem{zhang2016tweet}
M.~Zhang, Y.~Zhang, and G.~Fu, ``Tweet sarcasm detection using deep neural
  network,'' in \emph{Proceedings of COLING 2016, The 26th International
  Conference on Computational Linguistics: Technical Papers}, 2016, pp.
  2449--2460.

\bibitem{ptavcek2014sarcasm}
T.~Pt{\'a}{\v{c}}ek, I.~Habernal, and J.~Hong, ``Sarcasm detection on czech and
  english twitter,'' in \emph{Proceedings of COLING 2014, the 25th
  International Conference on Computational Linguistics: Technical Papers},
  2014, pp. 213--223.

\bibitem{wang2015twitter}
Z.~Wang, Z.~Wu, R.~Wang, and Y.~Ren, ``Twitter sarcasm detection exploiting a
  context-based model,'' in \emph{international conference on web information
  systems engineering}.\hskip 1em plus 0.5em minus 0.4em\relax Springer, 2015,
  pp. 77--91.

\bibitem{joshi2016harnessing}
A.~Joshi, V.~Tripathi, P.~Bhattacharyya, and M.~Carman, ``Harnessing sequence
  labeling for sarcasm detection in dialogue from tv series ‘friends’,'' in
  \emph{Proceedings of The 20th SIGNLL Conference on Computational Natural
  Language Learning}, 2016, pp. 146--155.

\bibitem{ghosh2016fracking}
A.~Ghosh and T.~Veale, ``Fracking sarcasm using neural network,'' in
  \emph{Proceedings of the 7th workshop on computational approaches to
  subjectivity, sentiment and social media analysis}, 2016, pp. 161--169.

\bibitem{vaswani2017attention}
A.~Vaswani, N.~Shazeer, N.~Parmar, J.~Uszkoreit, L.~Jones, A.~N. Gomez,
  {\L}.~Kaiser, and I.~Polosukhin, ``Attention is all you need,'' in
  \emph{Advances in neural information processing systems}, 2017, pp.
  5998--6008.

\bibitem{mikolov2013efficient}
T.~Mikolov, K.~Chen, G.~Corrado, and J.~Dean, ``Efficient estimation of word
  representations in vector space,'' \emph{arXiv preprint arXiv:1301.3781},
  2013.

\bibitem{mikolov2013distributed}
T.~Mikolov, I.~Sutskever, K.~Chen, G.~S. Corrado, and J.~Dean, ``Distributed
  representations of words and phrases and their compositionality,'' in
  \emph{Advances in neural information processing systems}, 2013, pp.
  3111--3119.

\bibitem{pennington2014glove}
J.~Pennington, R.~Socher, and C.~Manning, ``Glove: Global vectors for word
  representation,'' in \emph{Proceedings of the 2014 conference on EMNLP},
  2014, pp. 1532--1543.

\bibitem{joulin2017bag}
A.~Joulin, {\'E}.~Grave, P.~Bojanowski, and T.~Mikolov, ``Bag of tricks for
  efficient text classification,'' in \emph{Proceedings of the 15th Conference
  of the European Chapter of the ACL}, 2017, pp. 427--431.

\bibitem{peters2018deep}
M.~E. Peters, M.~Neumann, M.~Iyyer, M.~Gardner, C.~Clark, K.~Lee, and
  L.~Zettlemoyer, ``Deep contextualized word representations,'' in
  \emph{Proceedings of NAACL-HLT}, 2018, pp. 2227--2237.

\bibitem{devlin2019bert}
J.~Devlin, M.-W. Chang, K.~Lee, and K.~Toutanova, ``Bert: Pre-training of deep
  bidirectional transformers for language understanding,'' in \emph{Proceedings
  of the 2019 Conference of NAACL: Human Language Technologies}, 2019, pp.
  4171--4186.

\bibitem{zhou2016learning}
B.~Zhou, A.~Khosla, A.~Lapedriza, A.~Oliva, and A.~Torralba, ``Learning deep
  features for discriminative localization,'' in \emph{Proceedings of the IEEE
  conference on computer vision and pattern recognition}, 2016, pp. 2921--2929.

\bibitem{selvaraju2017grad}
R.~R. Selvaraju, M.~Cogswell, A.~Das, R.~Vedantam, D.~Parikh, and D.~Batra,
  ``Grad-cam: Visual explanations from deep networks via gradient-based
  localization,'' in \emph{Proceedings of the IEEE international conference on
  computer vision}, 2017, pp. 618--626.

\bibitem{oraby2016creating}
S.~Oraby, V.~Harrison, L.~Reed, E.~Hernandez, E.~Riloff, and M.~Walker,
  ``Creating and characterizing a diverse corpus of sarcasm in dialogue,'' in
  \emph{Proceedings of the 17th Annual Meeting of the Special Interest Group on
  Discourse and Dialogue}, 2016, pp. 31--41.

\bibitem{walker2012corpus}
M.~A. Walker, J.~E.~F. Tree, P.~Anand, R.~Abbott, and J.~King, ``A corpus for
  research on deliberation and debate.'' in \emph{LREC}.\hskip 1em plus 0.5em
  minus 0.4em\relax Istanbul, 2012, pp. 812--817.

\bibitem{khodak2018large}
M.~Khodak, N.~Saunshi, and K.~Vodrahalli, ``A large self-annotated corpus for
  sarcasm,'' in \emph{Proceedings of the Eleventh International Conference on
  Language Resources and Evaluation (LREC 2018)}, 2018.

\bibitem{misra2019sarcasm}
R.~Misra and P.~Arora, ``Sarcasm detection using hybrid neural network,''
  \emph{arXiv preprint arXiv:1908.07414}, 2019.

\bibitem{NEURIPS2019_9015}
A.~Paszke, S.~Gross, F.~Massa, A.~Lerer, J.~Bradbury, G.~Chanan, T.~Killeen,
  Z.~Lin, N.~Gimelshein, L.~Antiga, A.~Desmaison, A.~Kopf, E.~Yang, Z.~DeVito,
  M.~Raison, A.~Tejani, S.~Chilamkurthy, B.~Steiner, L.~Fang, J.~Bai, and
  S.~Chintala, ``Pytorch: An imperative style, high-performance deep learning
  library,'' in \emph{Advances in Neural Information Processing Systems 32},
  2019.

\bibitem{Wolf2019HuggingFacesTS}
T.~Wolf, L.~Debut, V.~Sanh, J.~Chaumond, C.~Delangue, A.~Moi, P.~Cistac,
  T.~Rault, R.~Louf, M.~Funtowicz, and J.~Brew, ``Huggingface's transformers:
  State-of-the-art natural language processing,'' \emph{ArXiv}, 2019.

\bibitem{farias2016irony}
D.~I.~H. Far{\'\i}as, V.~Patti, and P.~Rosso, ``Irony detection in twitter: The
  role of affective content,'' \emph{ACM Transactions on Internet Technology
  (TOIT)}, pp. 1--24, 2016.

\bibitem{tay2018reasoning}
Y.~Tay, A.~T. Luu, S.~C. Hui, and J.~Su, ``Reasoning with sarcasm by reading
  in-between,'' in \emph{Proceedings of the 56th Annual Meeting of the ACL},
  2018, pp. 1010--1020.

\bibitem{clark2019does}
K.~Clark, U.~Khandelwal, O.~Levy, and C.~D. Manning, ``What does bert look at?
  an analysis of bert’s attention,'' in \emph{Proceedings of the 2019 ACL
  Workshop BlackboxNLP: Analyzing and Interpreting Neural Networks for NLP},
  2019, pp. 276--286.

\end{thebibliography}

\end{document}